\documentclass{article}
\usepackage[preprint]{neurips_2026}

\usepackage[utf8]{inputenc}
\usepackage[T1]{fontenc}
\usepackage[hidelinks]{hyperref}
\usepackage{url}
\usepackage{booktabs}
\usepackage{amsfonts}
\usepackage{amsmath}
\usepackage{amssymb}
\usepackage{nicefrac}
\usepackage{microtype}
\usepackage{graphicx}
\usepackage{subcaption}
\usepackage{xcolor}
\usepackage{multirow}
\usepackage{algorithm}
\usepackage{algorithmic}

\usepackage{enumitem}
\usepackage{placeins}
\usepackage{threeparttable}
\usepackage{float}

\newcommand{\answerPartly}[1][]{\textcolor{olive}{[Partly]#1}}
\newcommand{\boldres}[1]{\textcolor{red}{\textbf{#1}}}
\newcommand{\secondres}[1]{\textcolor{blue}{\underline{#1}}}
\newcommand{\yhat}{\hat{y}}
\newcommand{\flin}{f_{\text{lin}}}
\newcommand{\fkan}{f_{\text{KAN}}}
\newcommand{\Ukan}{U_{\text{KAN}}}
\newcommand{\Snonlin}{S_{\text{nonlin}}}
\newcommand{\E}{\mathbb{E}}

\title{Temporal Functional Circuits: From Spline Plots to\\Faithful Explanations in KAN Forecasting}

\author{
  Naveen Mysore \\
  University of California, Santa Barbara \\
  Dyssonance.ai \\
  \texttt{naveenmysore@ucsb.edu}, \texttt{nmysore.work@gmail.com}
}

\begin{document}

\maketitle

\begin{abstract}
Unlike MLPs, Kolmogorov-Arnold Networks (KANs) expose explicit learnable
edge functions on every connection, enabling mechanistic explanation
in time-series forecasting. This paper introduces
\emph{Temporal Functional Circuits}, a framework that transforms KAN edge
functions from latent visualizations into faithful, temporally grounded
explanations. Built on a \emph{gated residual KAN} that decomposes forecasts
into a linear base and a sparsely activated KAN correction, the framework
(i) maps each edge to input lags via output-aware attribution, (ii) ranks
edges by learned activation range, and (iii) validates faithfulness through
edge-level interventions including zeroing and spline removal. Removing the learned B-spline component while retaining the base SiLU term
degrades forecasts, providing evidence that the spline shape itself carries
predictive value beyond the base activation. On four synthetic regimes of increasing complexity, the learned
gate opens progressively wider as signal complexity grows. On regime-switching signals, gated KAN achieves 59\% lower MSE than
linear-only models. Across
eight benchmarks, the gated architecture is competitive with
linear, attention, and MLP alternatives, while providing interpretable
edge functions that MLP-based corrections cannot offer.
\end{abstract}

\section{Introduction}
\label{sec:intro}

Long-term time-series forecasting (LTSF) is a core problem in climate science,
energy management, transportation, and healthcare~\citep{zhou2021informer,
wu2021autoformer}. Recent work has shown that simple linear models
can match or outperform complex Transformer architectures on standard
benchmarks~\citep{zeng2023dlinear}, raising the question of when additional
model complexity is justified.

Kolmogorov-Arnold Networks (KANs)~\citep{liu2024kan} provide an
alternative to MLPs: instead of fixed activation functions on nodes, KANs place
learnable univariate functions on edges, motivated by the Kolmogorov-Arnold
representation theorem~\citep{kolmogorov1957representation,
arnolddiscovery1963}. These edge functions, typically parameterized as
B-splines~\citep{deboor1978splines}, can be individually inspected, leading to
claims that KANs are ``inherently interpretable.'' Several recent works apply
KANs to time-series forecasting~\citep{vacarubio2024kan_tsf, xu2024kan4tsf,
qiu2024timekan}, citing this interpretability as a key advantage.

A recent investigation of KAN-based forecasting is consistent with this
complexity~\citep{anon2026decompkan}. In controlled ablation experiments
within a decomposition-patching pipeline, three forecasting cores (KAN,
linear, attention) were compared across standard benchmarks
(Table~\ref{tab:decompkan_ablation} in the appendix).
The result showed that \textbf{no single core dominates}: KAN ranked first
on Weather, linear on ETTh1, and results were near-tied on ETTm2. Each core ranked first on a
different subset of datasets. This finding motivated the central question
of this paper: \emph{if KAN is not universally better, when does it help,
and can we explain why?}

This paper argues that \textbf{visualizing learned splines is not the same as
explaining forecasts}. A KAN edge function $\phi_e(z)$ operates in latent
space: it does not directly indicate which input time steps or channels
activated it, whether the model actually used it (versus suppressing it via
downstream layers), or whether removing it would change the forecast. The
saliency-map literature has shown that visually plausible explanations can fail
basic sanity checks~\citep{adebayo2018sanity}, and time-series models pose
additional challenges because temporal autocorrelation can create superficially
meaningful but unfaithful attribution patterns.

This work proposes a framework that transforms KAN edge functions from
latent visualizations into \emph{faithful, temporally grounded explanations}.
The contributions are:

\begin{enumerate}[leftmargin=*,nosep]
\item \textbf{Gated Residual KAN.} A forecasting architecture
$\yhat = \flin(x) + g(x) \odot \fkan(x)$ with an $L_1$-regularized gate
$g(x) \in [0,1]^C$ that provides a measurable KAN utilization metric
$\Ukan$ (averaged over channels and branches; formal definition in
\S\ref{sec:model}).
\item \textbf{Temporal Functional Circuits.} An explanation tuple
$\mathcal{E}_e = (\phi_e, A_e, I_e, \Delta_e)$ per KAN edge,
comprising the learned function, lag attribution, importance,
and interventional effect (\S\ref{sec:tfc}).
\item \textbf{Faithfulness validation.} Synthetic mechanism recovery on four
regimes, edge deletion curves, and spline-removal interventions providing evidence
that the B-spline component accounts for up to 41\% of the full
zeroing effect on Weather (\S\ref{sec:experiments}).
\item \textbf{Empirical analysis.} Across eight benchmarks, the gated
architecture is competitive with linear, KAN-only, attention, and MLP
alternatives. Gate utilization correlates with an independent nonlinear
residual diagnostic ($\rho = 0.46$, $p = 0.015$;
\S\ref{sec:analysis}).
\end{enumerate}

\section{Related Work}
\label{sec:related}

\paragraph{KANs and KAN-based time-series forecasting.}
KANs~\citep{liu2024kan} place learnable univariate functions on edges rather
than fixed activations on nodes, motivated by the Kolmogorov-Arnold
representation theorem~\citep{kolmogorov1957representation,
arnolddiscovery1963}. Several works apply KANs to time-series forecasting:
\citet{vacarubio2024kan_tsf} first demonstrate KANs as temporal approximators;
KAN4TSF~\citep{xu2024kan4tsf} benchmarks KANs as MLP replacements;
TimeKAN~\citep{qiu2024timekan} integrates KANs with frequency decomposition;
and KAN4TSF/RMoK~\citep{xu2024kan4tsf} uses mixture-of-KAN-expert routing.
These works largely treat visualizable edge functions as interpretability
evidence. This paper argues that time-series forecasting requires \emph{temporal
grounding} and \emph{intervention-tested faithfulness}: the proposed gated residual
design isolates nonlinear correction (unlike RMoK's expert routing), and
Temporal Functional Circuits map edges to raw lags, rank them by
learned importance, and validate them through interventions.

\paragraph{Explainability for time series.}
Time-series XAI spans saliency-based
methods~\citep{ismail2020benchmarking, rojat2021survey, theissler2022review},
learned perturbation masks~\citep{crabbe2021dynamask}, windowed feature
importance~\citep{leung2023winit}, and contrastive surrogate
explanations~\citep{queen2023timex}. \citet{ismail2020benchmarking} benchmark
attribution methods on synthetic time series with known ground truth; this
evaluation paradigm is adopted for the synthetic mechanism-recovery experiments.
DynaMask~\citep{crabbe2021dynamask} learns dynamic masks over features and
time steps, providing a baseline for the proposed edge-to-lag attribution.
TimeX~\citep{queen2023timex} trains an interpretable surrogate to explain
black-box models. The present approach differs: rather than post-hoc surrogates,
the model is explained through its \emph{intrinsic} KAN edge functions, explicit
univariate functions that can be inspected, edited, and intervened upon.
Time-series shapelets~\citep{ye2009shapelets, grabocka2014learning} offer
another form of inherent interpretability via discriminative subsequences;
the proposed temporal circuits are analogous but operate on learned edge functions
grounded in forecast-horizon attribution rather than pattern matching.

\paragraph{Faithfulness and intervention-based evaluation.}
A central lesson from XAI research is that visual plausibility does not imply
faithfulness~\citep{adebayo2018sanity, jacovi2020faithfulness}.
\citet{adebayo2018sanity} showed that saliency maps can appear reasonable even
for randomly parameterized models. The attention-as-explanation
debate~\citep{jain2019attention, wiegreffe2019attention} further illustrates
that inspectable internal objects (attention weights, or by analogy KAN
splines) are not automatically faithful. \citet{rudin2019stop} argues for
intrinsically interpretable models; KANs partially satisfy this via explicit
edge functions, but this work shows that faithfulness still requires intervention
testing. The evaluation draws on attribution axioms from Integrated
Gradients~\citep{sundararajan2017axiomatic}, concept-level testing from
TCAV~\citep{kim2018tcav}, and the mechanistic-interpretability tradition of
activation patching and circuit analysis~\citep{olah2020zoom,
geiger2021causal}. These ideas are adapted to the time-series KAN setting: edge
removal and spline-removal interventions establish
\emph{model-internal} interventional effects, without claiming real-world
causality in the data-generating process.

\paragraph{Forecasting backbones and gating.}
The controlled comparison builds on DLinear~\citep{zeng2023dlinear},
PatchTST~\citep{nie2023patchtst}, iTransformer~\citep{liu2024itransformer},
and RevIN~\citep{kim2022revin}. The gated architecture relates to
MoE designs~\citep{jacobs1991adaptive, shazeer2017moe}, TFT's
gates~\citep{lim2021tft}, and MoLE~\citep{ni2024mole}. Recent KAN-specific
gating~\citep{sigkan2024, grkanmoe2024, moekan2024, hahnkan2026} uses
gating for \emph{performance routing}; our gate instead serves a dual
purpose: suppressing KAN when unnecessary \emph{and} providing a
measurable diagnostic ($\Ukan$). For edge importance, the original
KAN~\citep{liu2024kan} uses data-dependent $L_1$ pruning;
CausalKANs~\citep{causalkan2024} and QuantKAN~\citep{quantkan2025}
extend this to causal and quantized settings. The proposed $R_e$ instead
uses data-independent activation range over the B-spline grid.

\section{Gated Residual KAN}
\label{sec:model}

Given an input window $x \in \mathbb{R}^{L \times C}$ of $L$ time steps
across $C$ variates, the forecast $\yhat \in \mathbb{R}^{H \times C}$ is computed via:
\begin{equation}
\label{eq:gated}
\yhat = \flin(x) + g_{\text{trend}}(x) \odot \fkan^{\text{trend}}(x)
       + g_{\text{resid}}(x) \odot \fkan^{\text{resid}}(x),
\end{equation}
where $\flin$ is a linear forecasting branch, $\fkan^{\text{trend}}$ and
$\fkan^{\text{resid}}$ are KAN-based nonlinear branches applied to trend
and residual components respectively, and
$g_{\text{trend}}, g_{\text{resid}} \in [0,1]^C$ are per-channel scalar gates.

\paragraph{Pipeline.}
Both branches share a common preprocessing pipeline.
\textbf{(1)~RevIN}~\citep{kim2022revin} normalizes each window to zero mean
and unit variance per channel, storing statistics for denormalization.
\textbf{(2)~Adaptive normalization} encodes the temporal shape of the
RevIN-normalized input via a two-layer MLP into a compact statistics vector
$s \in \mathbb{R}^{d_s}$ ($d_s{=}8$), then produces a learned per-channel
scale and shift: $x' = \gamma(s) \cdot x + \beta(s)$, where
$(\gamma, \beta) = \text{Linear}(s)$ initialized to identity.
\textbf{(3)~Decomposition} splits the signal into trend and residual
components via a moving-average filter~\citep{wu2021autoformer}.
Each component is processed independently with its own gate
(Equation~\ref{eq:gated}). The linear branch combines both:
$\flin = \flin^{\text{trend}} + \flin^{\text{resid}}$.
All processing is channel-independent: $x$ is reshaped
from $[B, L, C]$ to $[B{\cdot}C, L]$ before entering the branches.

\paragraph{Linear branch.}
The input is divided into overlapping patches of length $P$ with stride $S$,
embedded via a linear layer into dimension $d$, flattened, and projected to
the forecast horizon $H$ through a single linear head. Under the shared
preprocessing pipeline, this branch captures temporal patterns expressible
as a linear function of patch embeddings.

\paragraph{KAN branch.}
The same patching produces token embeddings that are flattened and processed
through a B-spline KAN~\citep{liu2024kan, efficientkan2024} with two hidden
layers. Each KAN edge implements a learnable univariate function
$\phi_e(z) = w \cdot \text{SiLU}(z) + \sum_k c_k B_k^p(z)$,
where $B_k^p$ are cubic B-spline basis functions. This branch captures
nonlinear residual structure that the linear branch cannot.

\paragraph{Gate.}
The gate network maps the flattened input to a single scalar per channel
via a two-layer MLP with sigmoid activation:
$g_c(x) = \sigma(\text{MLP}(x_c)) \in [0,1]$.
The gate value is broadcast across all forecast horizons.

\paragraph{Training objective.}
\begin{equation}
\label{eq:loss}
\mathcal{L} = \text{MSE}(\yhat, y) + \lambda_g \left(
\E[\|g_{\text{trend}}(x)\|_1] + \E[\|g_{\text{resid}}(x)\|_1] \right),
\end{equation}
where $\lambda_g$ controls gate sparsity. The $L_1$ penalty encourages both
gates to remain closed (use linear only) unless the KAN correction genuinely
reduces forecast error.

\paragraph{KAN utilization.}
Since the architecture has separate gates for trend and residual branches,
we define $\Ukan = \frac{1}{2C}\sum_{c=1}^{C}(\E[g_{\text{trend},c}(x)] +
\E[g_{\text{resid},c}(x)])$ as a scalar summary (averaged over channels $C$) of how much each dataset
relies on nonlinear correction. A dataset with $\Ukan \approx 0$ is
well-served by linear forecasting; $\Ukan > 0$ suggests the trained model uses nonlinear KAN correction
under the chosen gate penalty.

\section{Temporal Functional Circuits}
\label{sec:tfc}

\begin{figure}[t]
\centering
\includegraphics[width=\textwidth]{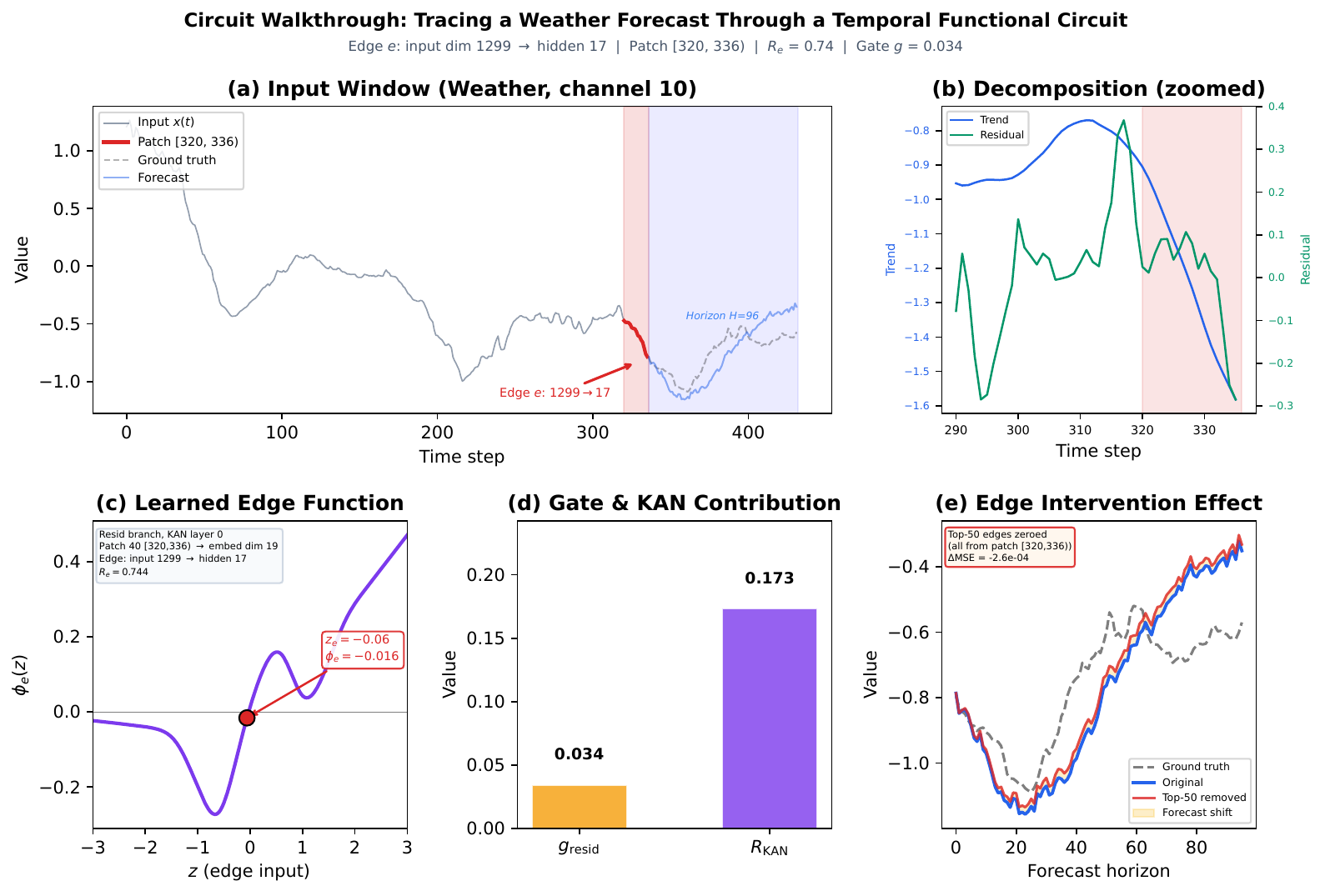}
\caption{Circuit walkthrough on Weather (channel 10). (a)~Input window
with the top-attributed patch [320, 336) highlighted. (b)~Trend/residual
decomposition at the patch region (dual axes; residual is small).
(c)~Learned B-spline edge function $\phi_e(z)$ for the top edge by
$R_e$, with the actual data point marked. (d)~Gate value and calibrated
KAN contribution $R_{\text{KAN}}$ for this window. (e)~Forecast with
and without the top-50 edges by $R_e$: removing them visibly shifts the
prediction, providing evidence that these edges carry functional load.}
\label{fig:walkthrough}
\end{figure}

A KAN edge function $\phi_e(z)$ operates in latent space; it does not
directly reveal which input lags drove it, whether the model actually used it,
or whether removing it changes the forecast.
A \emph{Temporal Functional Circuit} is defined as a tuple
$\mathcal{E}_e = (\phi_e, A_e, I_e, \Delta_e)$ that answers each of
these questions. Figure~\ref{fig:walkthrough} illustrates a complete
circuit walkthrough on a Weather test window: from the input signal
through decomposition, to the learned edge function and its
interventional effect on the forecast.

\subsection{Edge-to-Lag Attribution}
\label{sec:attribution}

Each KAN edge $e$ in the first layer maps from patched input dimension $i$ to
hidden dimension $j$. Through the patch embedding geometry, dimension $i$
corresponds to patch index $p = \lfloor i/d \rfloor$, providing a coarse
receptive field over raw lags $[p \cdot S, \, p \cdot S + P)$. Because the
patch embedding mixes all $P$ lag values through a learned linear projection,
exact lag attribution requires gradients through the embedding. The
output-aware edge contribution:
\begin{equation}
C_e(x) = \phi_e(z_e(x)) \cdot \frac{\partial \yhat}{\partial a_e},
\end{equation}
where $a_e = \phi_e(z_e)$ is the scalar edge activation and
$\partial \yhat / \partial a_e$ captures downstream sensitivity (summed
over output channels and horizons for scalarization, with stop-gradient
to avoid second-order derivatives). Because $\yhat$ depends on $g(x) \odot f_{\text{KAN}}(x)$ (Equation~\ref{eq:gated}), the gate is implicitly incorporated in $\partial \yhat / \partial a_e$ via the chain rule; when the gate is closed, the gradient is near zero regardless of edge activation. The temporal attribution is:
\begin{equation}
\label{eq:attribution}
A_e(c, t) = \E_{x \sim \mathcal{D}} \left[
\left| \frac{\partial C_e(x)}{\partial x_{c,t}} \right|
\right].
\end{equation}
Because the model is channel-independent ($x$ is reshaped to $[B{\cdot}C, L]$),
$c$ indexes the channel \emph{being processed}, not cross-channel influence.
$A_e(c,t)$ tells us which time steps within each channel the edge responds to.

\subsection{Data-Weighted Edge Importance}
\label{sec:importance}

Edges are ranked by their expected contribution to the forecast:
\begin{equation}
\label{eq:importance}
I_e = \E_{x \sim \mathcal{D}} \left[
\left| \phi_e(z_e(x)) \cdot \frac{\partial \yhat}{\partial a_e}
\right|
\right].
\end{equation}
This combines two observable factors: (1) edge output magnitude---does the edge fire on real data? and (2) downstream sensitivity---does the edge affect the forecast? The gate's influence enters implicitly through the gradient: $\partial \yhat / \partial a_e = g(x) \cdot \partial f_{\text{KAN}} / \partial a_e \cdot (\partial \yhat / \partial f_{\text{KAN}})$, so edges in a closed-gate branch automatically receive near-zero importance. Edges with
large $I_e$ are the functionally important components of the nonlinear correction.

A simpler alternative ranks edges by their activation range
$R_e = \max_{z \in [t_0, t_G]} \phi_e(z) - \min_{z \in [t_0, t_G]} \phi_e(z)$,
evaluated over the B-spline grid interval $[t_0, t_G]$ (the learned knot
boundaries), which measures functional capacity without requiring data. Direct comparison shows $R_e$
produces 20$\times$ larger deletion effects than $I_e$ on Weather
(Appendix~\ref{app:temporal_grounding}), suggesting that for B-spline KANs,
learned functional capacity is itself a strong faithfulness signal. $R_e$
is therefore used as the primary ranking in deletion experiments.

\subsection{Edge Interventions}
\label{sec:interventions}

To validate faithfulness, the \emph{interventional effect} of
editing edge $e$ on predictive loss:
\begin{equation}
\label{eq:intervention}
\Delta_e = \E_{(x,y) \sim \mathcal{D}_{\text{test}}} \left[
\ell(f^{(-e)}(x), y) - \ell(f(x), y)
\right],
\end{equation}
where $f^{(-e)}$ denotes the model with edge $e$ intervened upon.
Positive $\Delta_e$ means the edge is beneficial. Three intervention
types are considered:
\begin{itemize}
\item \textbf{Zero}: $a_e(x) \leftarrow 0$ (silence the edge completely).
\item \textbf{Mean}: $a_e(x) \leftarrow \E[a_e]$ (replace with population
mean, avoiding out-of-distribution internal states).
\item \textbf{Spline removal}: remove only the B-spline component, retaining
$\phi_e(z) = w \cdot \text{SiLU}(z)$ (tests whether the learned spline shape
matters beyond the base activation; this is KAN-specific).
\end{itemize}

\emph{Note:} $\Delta_e$ measures the edge's interventional effect on the
\emph{model's computation}, not a causal claim about the data-generating
process.

\paragraph{Deletion curves.}
For aggregate faithfulness, all edges are ranked by an importance proxy
(computed on validation data) and progressively delete the top-$k$, random-$k$,
or bottom-$k$ edges, measuring MSE increase on the \emph{test set}. A faithful
ranking satisfies:
$\Delta_{\text{top-}k} > \Delta_{\text{random-}k} > \Delta_{\text{bottom-}k}$.

\section{Experiments}
\label{sec:experiments}

\subsection{Setup}
\label{sec:setup}

\paragraph{Datasets.}
The evaluation covers eight benchmarks: seven standard LTSF datasets
(Weather, 21 variates; Solar, 137; ECL, 321; ETTh1, ETTh2, ETTm1, ETTm2,
7 each) plus PPG-DaLiA~\citep{reiss2019ppg} (12 variates, physiological
signals).
All datasets use chronological train/validation/test splits following
established LTSF protocols. ECL is included in the core ablation
(Table~\ref{tab:killrisk}) but excluded from multi-horizon and
diagnostic analyses due to its high computational cost (321 variates
$\times$ 4 horizons $\times$ 3 seeds); all reported correlations use
$n=28$ (7 datasets $\times$ 4 horizons).

\paragraph{Architecture.}
All cores share an identical pipeline: RevIN, adaptive normalization,
moving-average decomposition ($K=25$), and patching ($P=16$, $S=8$,
$d=32$). The input look-back window is $L=336$ for most datasets
and $L=512$ for ECL and ETTm1. KAN branches use grid size $G=5$, spline order $p=3$, hidden
dimension 64, depth 2. The \emph{only} variable across ablation rows is
the core type.

\paragraph{Training.}
Adam optimizer (initial learning rate $10^{-3}$ for Weather/ETTh2/ETTm2, $2{\times}10^{-4}$ for ETTh1/ETTm1/Solar; see Appendix~\ref{app:hyperparams}), batch size 256 (Solar: 512, ECL: 128), cosine learning rate schedule, 50 epochs, patience 10.
Gate regularization $\lambda_g = 0.05$ for all datasets except Solar
($\lambda_g = 0.01$, selected via validation; see
Appendix~\ref{app:gate_ablation} for ablation).
Table~\ref{tab:killrisk} reports mean $\pm$ std over 3 seeds
(42, 123, 456). Synthetic experiments (Table~\ref{tab:synthetic}) also
report 3-seed statistics.

\subsection{Core Ablation}
\label{sec:killrisk}

Table~\ref{tab:killrisk} compares six architectures within the same pipeline
at forecast horizon $H=96$. The key question is whether KAN's spline-based
nonlinearity provides unique value.

\begin{table}[t]
\centering
\caption{Core ablation at $H=96$ (MSE, mean$\pm$std over 3 seeds).
Bold: best unrounded mean per dataset; rounded ties may appear identical.
All cores use identical preprocessing.}
\label{tab:killrisk}
\begin{scriptsize}
\setlength{\tabcolsep}{3pt}
\begin{tabular}{lcccccccc}
\toprule
Core & Weather & ETTh1 & ETTh2 & ETTm1 & ETTm2 & Solar & ECL & PPG \\
\midrule
Linear & .152{\tiny$\pm$.001} & .453{\tiny$\pm$.010} & .240{\tiny$\pm$.002} & .350{\tiny$\pm$.002} & .150{\tiny$\pm$.002} & .204{\tiny$\pm$.001} & .131{\tiny$\pm$.000} & .564{\tiny$\pm$.001} \\
KAN & \textbf{.149}{\tiny$\pm$.001} & .477{\tiny$\pm$.022} & .244{\tiny$\pm$.000} & .348{\tiny$\pm$.000} & .150{\tiny$\pm$.001} & .183{\tiny$\pm$.000} & .132{\tiny$\pm$.000} & \textbf{.559}{\tiny$\pm$.001} \\
Attention & .150{\tiny$\pm$.002} & .460{\tiny$\pm$.012} & .251{\tiny$\pm$.003} & .347{\tiny$\pm$.002} & .150{\tiny$\pm$.001} & .180 & \textbf{.129}{\tiny$\pm$.000} & .564{\tiny$\pm$.002} \\
Ungated & .150{\tiny$\pm$.001} & \textbf{.451}{\tiny$\pm$.006} & .245{\tiny$\pm$.003} & .347{\tiny$\pm$.002} & .149{\tiny$\pm$.001} & \textbf{.180}{\tiny$\pm$.001} & .132{\tiny$\pm$.000} & .563{\tiny$\pm$.002} \\
\textbf{Gated KAN} & .149{\tiny$\pm$.001} & \textbf{.451}{\tiny$\pm$.007} & .243{\tiny$\pm$.003} & \textbf{.345}{\tiny$\pm$.003} & \textbf{.148}{\tiny$\pm$.001} & .182{\tiny$\pm$.000} & .132{\tiny$\pm$.001} & .564{\tiny$\pm$.001} \\
Gated MLP & .150{\tiny$\pm$.001} & .453{\tiny$\pm$.008} & \textbf{.239}{\tiny$\pm$.003} & .347{\tiny$\pm$.003} & .149{\tiny$\pm$.001} & .185{\tiny$\pm$.001} & .131{\tiny$\pm$.000} & .563{\tiny$\pm$.001} \\
\bottomrule
\end{tabular}
\end{scriptsize}

\vspace{0.5em}
{\footnotesize Solar uses $\lambda_g=0.01$; all others $\lambda_g=0.05$.
Attention Solar: single seed. ETTh1 has high seed variance ($\sigma \approx 0.01$--$0.02$).}
\end{table}

\paragraph{Findings.}
\textbf{No single core dominates}: Gated KAN is best or tied-best on
4/8 datasets (Weather, ETTh1, ETTm1, ETTm2), but Gated MLP wins on
ETTh2 and many differences are within one standard deviation.
Gated KAN improves over Ungated KAN on 4/8 datasets; the main value of
gating is diagnostic control over nonlinear correction rather than
uniform accuracy improvement.

KAN's value is therefore \emph{interpretability}: Gated MLP can match
accuracy, but cannot provide explicit edge functions for temporal circuits.
Baselines and basis ablations are in
Appendices~\ref{app:baselines}--\ref{app:basis}.

\subsection{KAN Utilization}
\label{sec:utilization}

The gate produces a measurable diagnostic $\Ukan$ per dataset
(full table in Appendix~\ref{app:horizons}).
At $H$=96, $\Ukan$ ranges from 0.022 (PPG, ETTh1---predominantly linear)
to 0.144 (Solar---highest nonlinear demand). ETTh1 and ETTm2
show increasing $\Ukan$ at long horizons (0.022$\to$0.288,
0.020$\to$0.198), suggesting nonlinear correction becomes more
important for difficult far-horizon forecasts.
$\Ukan$ reflects both nonlinear demand and $L_1$ penalty; the calibrated
$R_{\text{KAN}}$ in Section~\ref{sec:analysis} controls for branch scaling.

\subsection{Synthetic Mechanism Recovery}
\label{sec:synthetic}

To validate that the gate and KAN respond to genuine signal complexity,
four synthetic univariate regimes with increasing complexity are constructed:
\begin{itemize}
\item \textbf{Linear}: $x(t) = \sin(2\pi \cdot 0.02\, t) + \epsilon$.
\item \textbf{Multi-frequency periodic}: $x(t) = \sin(2\pi \cdot 0.05\, t)
  + 0.8\,\text{tri}(2\pi \cdot 0.02\, t) + 0.5\sin(2\pi \cdot 0.13\, t)
  + \epsilon$, where $\text{tri}$ is a triangular wave (piecewise linear,
  requiring many AR terms).
\item \textbf{Threshold AR (SETAR)}: state-dependent dynamics---$x(t) = 0.6\,x(t{-}1) - 0.3\,x(t{-}2)
  + \epsilon$ if $x(t{-}1) > 0$, else $-0.4\,x(t{-}1) + 0.5\,x(t{-}2)
  + \epsilon$. Provably nonlinear (no linear AR can represent
  state-dependent coefficient switching).
\item \textbf{Regime-switching}: alternates between linear and multi-frequency
regimes every 500 steps with hidden state $z_t \in \{0,1\}$.
\end{itemize}

\begin{figure}[t]
\centering
\includegraphics[width=\textwidth]{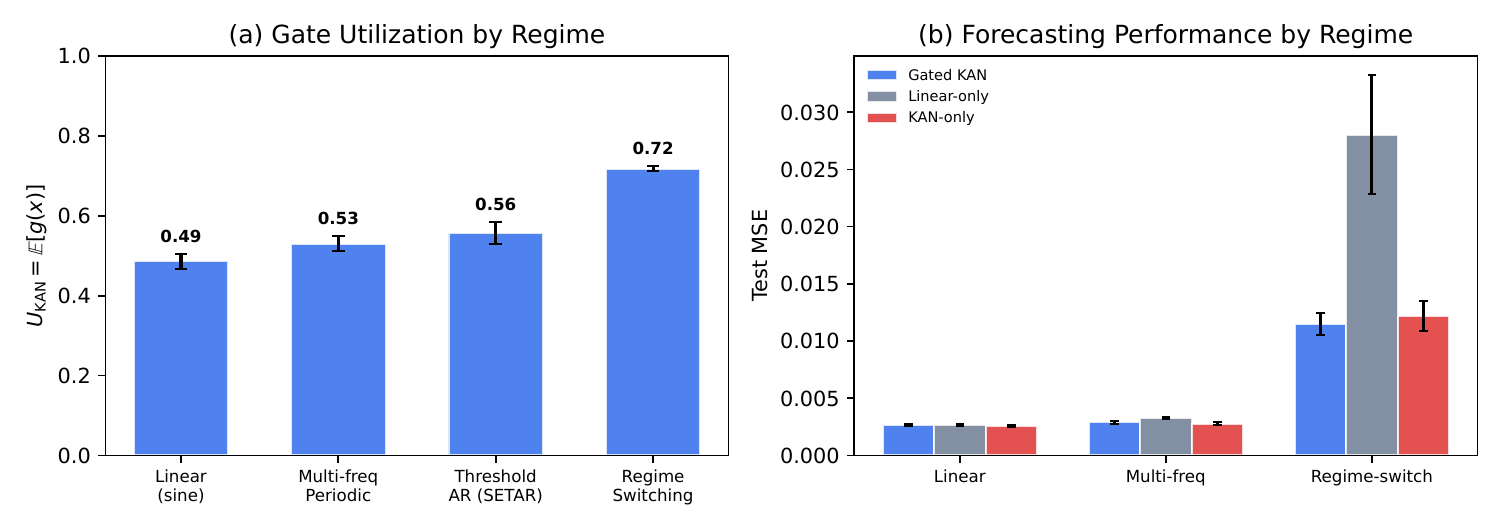}
\caption{Synthetic regime experiments. (a)~Gate utilization $\Ukan$ increases
monotonically with signal complexity: 0.49 (linear) $\to$ 0.53
(multi-frequency) $\to$ 0.56 (threshold AR) $\to$ 0.72
(regime-switching). (b)~On regime-switching, Gated KAN matches KAN-only
while outperforming Linear-only by 59\%.}
\label{fig:synthetic}
\end{figure}

Table~\ref{tab:synthetic} reports results averaged over 3 seeds.
Figure~\ref{fig:synthetic} visualizes the key findings.

\begin{table}[t]
\centering
\caption{Synthetic regime experiments. MSE (mean$\pm$std over 3
seeds) and KAN utilization $\Ukan$. Gate regularization $\lambda_g = 0$
(no sparsity pressure), testing monotonic gate response to nonlinear
complexity rather than sparse utilization.}
\label{tab:synthetic}
\small
\begin{tabular}{llccc}
\toprule
Regime & Core & MSE & $\Ukan$ \\
\midrule
\multirow{3}{*}{Linear (sine)}
& Gated KAN & .0027$\pm$.0001 & 0.49 \\
& Linear-only & .0027$\pm$.0001 & -- \\
& KAN-only & .0026$\pm$.0001 & -- \\
\midrule
\multirow{3}{*}{Multi-freq periodic}
& Gated KAN & .0029$\pm$.0001 & 0.53 \\
& Linear-only & .0033$\pm$.0001 & -- \\
& KAN-only & \textbf{.0028}$\pm$.0001 & -- \\
\midrule
\multirow{3}{*}{Threshold AR}
& Gated KAN & .990$\pm$.034 & 0.56 \\
& Linear-only & .986$\pm$.034 & -- \\
& KAN-only & \textbf{.970}$\pm$.030 & -- \\
\midrule
\multirow{3}{*}{Regime-switching}
& Gated KAN & \textbf{.0115}$\pm$.0009 & 0.72 \\
& Linear-only & .0280$\pm$.0052 & -- \\
& KAN-only & .0122$\pm$.0013 & -- \\
\bottomrule
\end{tabular}
\end{table}

\paragraph{Findings.}
(1)~\textbf{Gate utilization tracks signal complexity}: $\Ukan$ increases
monotonically (0.49$\to$0.53$\to$0.56$\to$0.72) without $L_1$ pressure.
The calibrated $R_{\text{KAN}}$ is even more discriminative: 0.05 (linear)
$\to$0.13 (multi-freq) $\to$0.50 (threshold AR) $\to$0.42
(regime-switching), showing the KAN branch contributes 5\% vs.\ 50\% of
the forecast on linear vs.\ nonlinear data (Appendix~\ref{app:temporal_grounding}).
(2)~\textbf{KAN correction reduces error on complex regimes}: on
regime-switching, gated KAN (MSE 0.012) and KAN-only (0.012) both
substantially outperform linear-only (0.028), a 59\% MSE reduction.
The threshold AR regime (a provably nonlinear SETAR process with
state-dependent dynamics) is difficult for all models but still elicits
higher gate opening. On the simple linear regime, all cores
perform identically; KAN does not hurt.
(3)~\textbf{Gated KAN matches KAN-only}: across all regimes, gated KAN
achieves MSE comparable to KAN-only, showing the gate does not sacrifice
accuracy.

\subsection{Edge Faithfulness}
\label{sec:faithfulness}

Edge importance rankings are validated by testing whether removal of
high-ranked edges degrades the forecast. All first-layer KAN edges are ranked
by activation range $R_e = \max_z \phi_e(z) - \min_z \phi_e(z)$, which
measures the learned functional capacity of each edge. The top-$k$,
random-$k$ (averaged over 50 draws), or bottom-$k$ edges are then deleted
by zeroing their weights, and MSE change is measured on the held-out test
set.

\begin{table}[t]
\centering
\caption{Edge deletion on the residual branch. Values are
$\Delta\text{MSE} \times 10^{4}$. Edges ranked by activation range $R_e$.
Weather random reports $\pm$1 std over 50 draws; other datasets report
mean only.}
\label{tab:faithful}
\small
\begin{tabular}{lcrrr}
\toprule
Dataset & $k$ & $\Delta_{\text{top}}$ & $\Delta_{\text{random}}$ & $\Delta_{\text{bottom}}$ \\
\midrule
\multirow{3}{*}{Weather}
& 5  & +0.14 & 0.00$\pm$0.01 & 0.00 \\
& 10 & +0.44 & 0.00$\pm$0.02 & 0.00 \\
& 50 & +2.20 & $-$0.01$\pm$0.05 & 0.00 \\
\midrule
\multirow{2}{*}{ETTm2}
& 10 & +0.02 & 0.00 & 0.00 \\
& 50 & +0.20 & 0.00 & $<$0.01 \\
\midrule
\multirow{2}{*}{PPG}
& 5  & +0.19 & 0.00 & 0.00 \\
& 10 & +0.40 & 0.00 & 0.00 \\
\midrule
\multirow{2}{*}{Solar}
& 5  & +0.06 & 0.00 & 0.00 \\
& 50 & +1.44 & +0.06 & $<$0.01 \\
\bottomrule
\end{tabular}
\end{table}

\paragraph{Findings.}
Across Weather, ETTm2, PPG, and Solar, deletion of top-ranked edges
consistently produces larger MSE increases than random or bottom-ranked
deletion (Weather random: 50 draws, $\pm$1 std reported). On Weather, top-50 deletion increases MSE by $2.20 \times 10^{-4}$
(0.15\% relative), while random deletion is consistent with zero
($-0.01 \pm 0.05 \times 10^{-4}$, 50 draws). On datasets where $\Ukan$ is very low
(ETTh1, ETTh2), deletion effects are near zero because the model barely
uses the KAN branch.

This provides evidence that activation range identifies edges whose removal
causes consistent, non-random forecast degradation. The actual learned
edge functions $\phi_e(z)$ for the top-ranked edges are visualized in
Appendix~\ref{app:edge_vis}.

\paragraph{Intervention types.}
Two interventions are tested on the top-20 residual edges (by $R_e$):
zeroing and spline removal (removing only the B-spline component,
retaining the base $\text{SiLU}$ term). On Weather, spline removal accounts for 41\% of the full zeroing effect
($\Delta$MSE $+3.8 \times 10^{-5}$ vs.\ $+9.1 \times 10^{-5}$). On
ETTh1 (low $\Ukan$), the fraction drops to 6\%. This test is
KAN-specific: it isolates the value of the learned spline shape, which
MLP connections cannot provide.

\paragraph{Temporal grounding validation.}
On a synthetic signal $x(t) = 0.6\, x(t{-}3) + 0.25\, x(t{-}12) + \epsilon$
with known causal lags, the proposed branch-level $A_e$ recovers true lags
with AUPRC = 0.44 ($6\times$ above random), while standard IG achieves
0.53. The key advantage of $A_e$ is edge-level resolution: it identifies
\emph{which KAN edges} implement each lag dependency, which model-level
methods cannot provide. Attributions are stable
across seeds (Pearson $r = 0.94$--$0.97$). IG-based input masking on
Weather increases MSE by 104\% (top-5) vs.\ 3\% (random), indicating the
model relies on specific temporal positions. A direct comparison of
edge-level deletion using $R_e$ vs.\ $I_e$ rankings shows $R_e$ produces
20$\times$ larger deletion effects on Weather, providing evidence that for
B-spline KANs, learned functional capacity is itself a faithful importance
signal (Appendix~\ref{app:temporal_grounding}).

\begin{figure}[t]
\centering
\includegraphics[width=\textwidth]{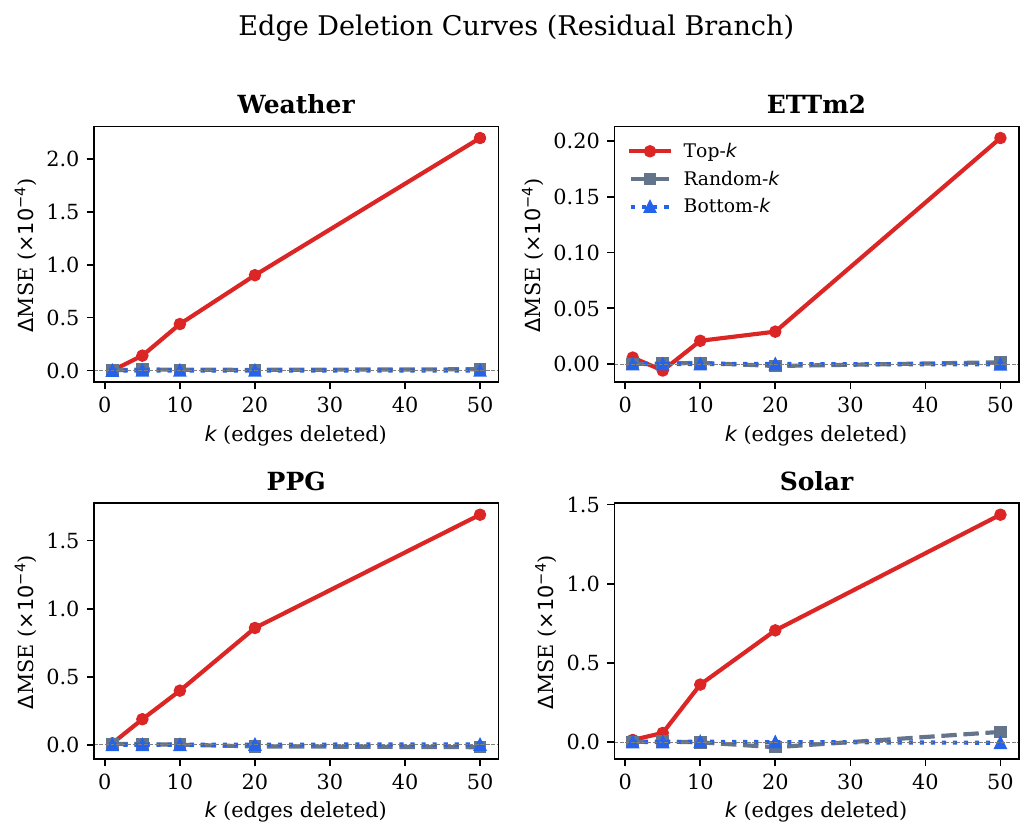}
\caption{Edge deletion curves (residual branch, ranked by $R_e$). Removing
top-ranked edges (red) consistently increases MSE more than random (gray)
or bottom-ranked (blue) deletion, providing evidence that activation-range ranking
identifies functionally relevant edges. The effect is strongest on Weather and Solar,
which have moderate-to-high $\Ukan$.}
\label{fig:deletion}
\end{figure}



\section{Analysis}
\label{sec:analysis}

\paragraph{Nonlinear residual diagnostic.}
An independent diagnostic $\Snonlin = (\text{MSE}_{\text{lin}} -
\text{MSE}_{\text{MLP}}) / \text{MSE}_{\text{lin}}$ is computed per
dataset-horizon pair, where $\text{MSE}_{\text{lin}}$ and
$\text{MSE}_{\text{MLP}}$ correspond to the Linear-only and Gated MLP
cores from Table~\ref{tab:killrisk} respectively. The Spearman correlation between $\Snonlin$ and
$\Ukan$ is $\rho = 0.46$ ($p = 0.015$, $n = 28$), consistent with the gate
tracking nonlinear residual structure that an independent diagnostic also
detects (scatter plot in Appendix~\ref{app:horizons}).

\paragraph{Calibrated KAN contribution.}
$\Ukan$ can be confounded by branch scaling. The calibrated metric
$R_{\text{KAN}} = \E[\|g(x) \fkan(x)\|_2 / \|\yhat(x)\|_2]$
controls for this: Weather has $R_{\text{KAN}} = 0.38$ (38\% of forecast from KAN),
ETTh1 $R_{\text{KAN}} = 0.004$ (predominantly linear), Solar
$R_{\text{KAN}} = 0.71$ (highest). Adaptive $\lambda_g$ selection
(Appendix~\ref{app:gate_ablation}) remains an open problem.

\paragraph{Why KAN over MLP.}
Gated MLP matches accuracy (Table~\ref{tab:killrisk}), but MLP weights
are opaque: no inspectable edge functions, no single-connection spline
removal, no individually editable functional units.
TFC exploits the explicit univariate functions KANs expose.

\section{Limitations}
\label{sec:limitations}

Edge interventions establish \emph{model-internal} effects, not causal
claims about the data-generating process.
Attribution is gradient-based and most direct for first-layer edges.
KAN edge functions are inspectable but not automatically human-semantic.
Gated KAN is competitive but not uniformly state-of-the-art;
stronger regime-adaptive gating remains future work.

\section{Conclusion}
\label{sec:conclusion}

This paper argued that KAN interpretability requires temporal grounding
and intervention testing, not spline visualization alone.
Temporal Functional Circuits $(\phi_e, A_e, I_e, \Delta_e)$ map edges to
input lags and validate faithfulness through interventions. Gate
utilization tracks synthetic signal complexity, and activation-range
deletion provides evidence that high-ranked edges are functionally
important.

\bibliography{references}
\bibliographystyle{plainnat}

\newpage
\appendix

\section{Motivating Investigation: DecompKAN}
\label{app:decompkan}

This work was motivated by an investigation of DecompKAN, an attention-free
forecasting architecture combining trend-residual decomposition, patching,
learned normalization, and B-spline KAN layers~\citep{anon2026decompkan}.
In controlled experiments (same training recipe across all models),
DecompKAN achieved competitive results, ranking first on 5 of 7 datasets
by MSE at $H$=96 (Table~\ref{tab:decompkan_ablation}).

\begin{figure}[h]
\centering
\includegraphics[width=\textwidth]{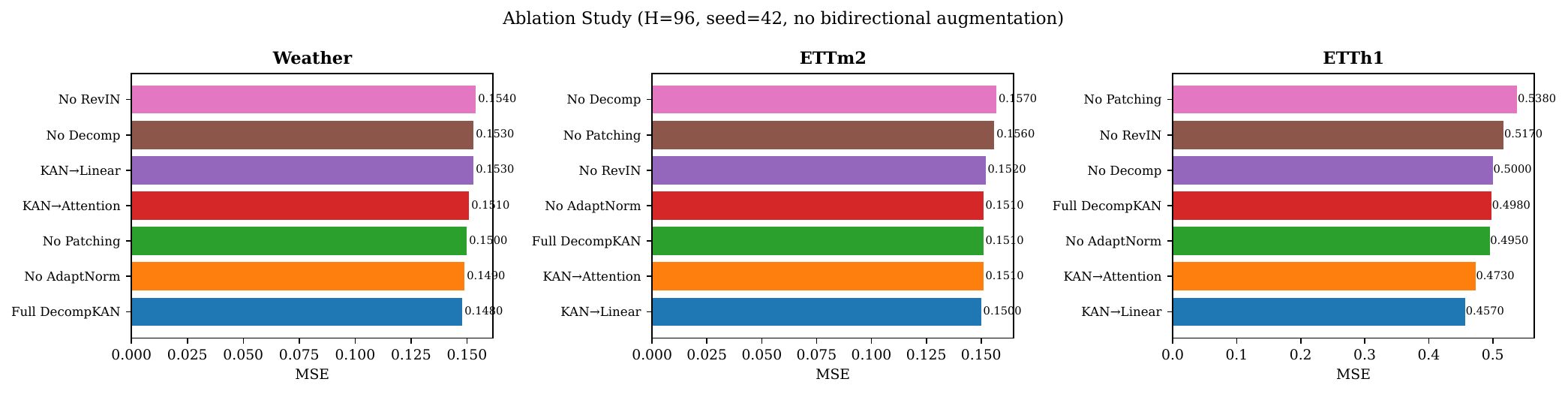}
\caption{DecompKAN ablation study ($H$=96). On Weather, the full KAN
pipeline is best. On ETTh1, replacing KAN with a linear layer improves
performance. The choice of nonlinear core is dataset-dependent, motivating
the gated residual design.}
\label{fig:decompkan_ablation}
\end{figure}

However, ablation analysis revealed that \textbf{the pipeline
(decomposition, patching, normalization) drives performance more than the
choice of nonlinear layer}. Figure~\ref{fig:decompkan_ablation} shows that
on ETTh1, replacing KAN with a linear layer actually \emph{improves}
performance, while on Weather, KAN provides genuine benefit
(see Table~\ref{tab:decompkan_ablation} for exact values). This dataset-dependent behavior, KAN helps on
some datasets but not others, is what led to the question: \emph{when does
KAN help, and can we explain why?}

The gated residual architecture (Section~\ref{sec:model}) directly addresses
this by learning \emph{when} to apply KAN correction, while Temporal
Functional Circuits (Section~\ref{sec:tfc}) explain \emph{what} the KAN
edges learn when they are used.

\begin{table}[!htbp]
\caption{DecompKAN controlled comparison ($H$=96, seed=42). All models
trained with the same recipe. \boldres{Red}: best. \secondres{Blue}: second.
DecompKAN ranks first on 5/7 datasets (MSE), but ablation
(Figure~\ref{fig:decompkan_ablation}) shows the pipeline matters more than
the KAN core, motivating the gated residual investigation.}
\label{tab:decompkan_ablation}
\vskip 0.02in
\centering
\begin{scriptsize}
\renewcommand{\arraystretch}{0.9}
\setlength{\tabcolsep}{4pt}
\begin{tabular}{l|cc|cc|cc|cc}
\toprule
\multirow{2}{*}{Dataset} &
\multicolumn{2}{c|}{\textbf{DecompKAN}} &
\multicolumn{2}{c|}{PatchTST} &
\multicolumn{2}{c|}{iTransformer} &
\multicolumn{2}{c}{DLinear} \\
& MSE & MAE & MSE & MAE & MSE & MAE & MSE & MAE \\
\midrule
Weather & \boldres{0.146} & \boldres{0.194} & \secondres{0.155} & \secondres{0.206} & 0.170 & 0.220 & 0.181 & 0.246 \\
Solar & \boldres{0.176} & \boldres{0.227} & \secondres{0.179} & \secondres{0.236} & 0.187 & 0.236 & 0.221 & 0.298 \\
ECL & \boldres{0.130} & \boldres{0.226} & 0.149 & 0.258 & \secondres{0.132} & \secondres{0.228} & 0.135 & 0.233 \\
ETTh1 & \secondres{0.450} & \secondres{0.453} & 0.478 & 0.467 & 0.474 & 0.465 & \boldres{0.444} & \boldres{0.442} \\
ETTh2 & \boldres{0.233} & \boldres{0.323} & 0.248 & 0.340 & 0.252 & 0.338 & \secondres{0.234} & \secondres{0.325} \\
ETTm1 & \boldres{0.349} & \boldres{0.384} & 0.382 & 0.400 & 0.370 & 0.399 & \secondres{0.363} & \secondres{0.391} \\
ETTm2 & \secondres{0.150} & \boldres{0.253} & 0.163 & 0.267 & 0.183 & 0.293 & \boldres{0.150} & \secondres{0.254} \\
\bottomrule
\end{tabular}
\end{scriptsize}
\end{table}

\section{Full Horizon Results}
\label{app:horizons}

\begin{table}[h]
\centering
\caption{Gated Residual KAN: MSE and $\Ukan$ across all horizons (seed=42).
$\Ukan$ reveals how gate utilization changes with forecast difficulty.}
\label{tab:full_horizons}
\small
\begin{tabular}{l|cccc|cccc}
\toprule
& \multicolumn{4}{c|}{Test MSE} & \multicolumn{4}{c}{$\Ukan$} \\
Dataset & 96 & 192 & 336 & 720 & 96 & 192 & 336 & 720 \\
\midrule
Weather & .150 & .193 & .242 & .327 & .046 & .044 & .041 & .036 \\
ETTh1 & .443 & .493 & .546 & .667 & .022 & .050 & .194 & .288 \\
ETTh2 & .245 & .288 & .347 & .451 & .044 & .050 & .110 & .058 \\
ETTm1 & .343 & .399 & .442 & .503 & .041 & .033 & .018 & .039 \\
ETTm2 & .147 & .190 & .230 & .293 & .020 & .049 & .074 & .198 \\
PPG & .565 & .620 & .719 & .945 & .022 & .017 & .019 & .021 \\
Solar & .182 & .197 & .205 & .208 & .144 & .139 & .153 & .151 \\
\bottomrule
\end{tabular}
\end{table}

Notable patterns: ETTh1 and ETTm2 show sharply increasing $\Ukan$ at longer
horizons (0.022$\to$0.288 and 0.020$\to$0.198), suggesting nonlinear
correction becomes more important for difficult far-horizon forecasts.
Solar maintains consistently high utilization ($\sim$0.14--0.15) across all
horizons. PPG remains uniformly low ($\sim$0.02), indicating predominantly
linear dynamics regardless of horizon.

\begin{figure}[h]
\centering
\includegraphics[width=0.65\textwidth]{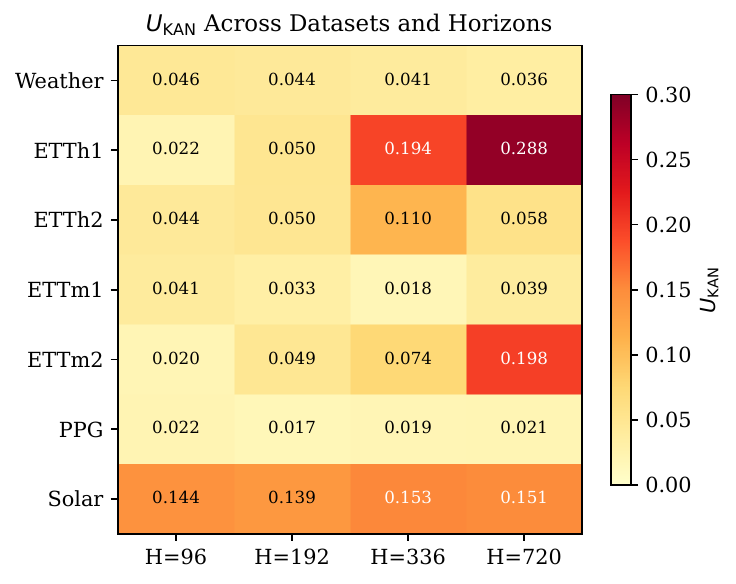}
\caption{$\Ukan$ heatmap across datasets and horizons.}
\end{figure}

\begin{figure}[h]
\centering
\includegraphics[width=0.65\textwidth]{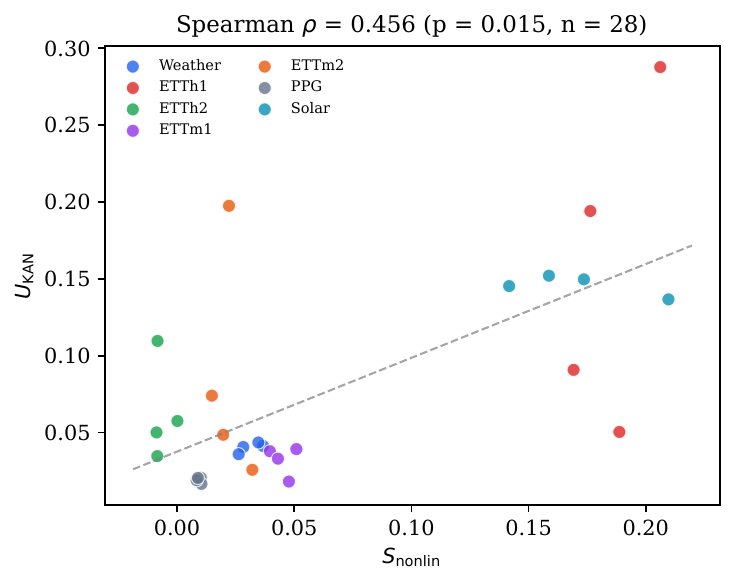}
\caption{$\Snonlin$ vs.\ $\Ukan$ scatter. Spearman $\rho = 0.46$, $p = 0.015$.}
\end{figure}

\FloatBarrier
\section{Published Baselines}
\label{app:baselines}

\begin{table}[h]
\centering
\caption{Architectural comparison under identical training conditions
($H$=96, seed=42). All models use the same preprocessing pipeline,
optimizer, schedule, and hyperparameters; only the forecasting core
differs. This isolates the effect of the core architecture rather than
training recipe.}
\label{tab:baselines}
\small
\begin{tabular}{lccccccc}
\toprule
Model & Weather & ETTh1 & ETTh2 & ETTm1 & ETTm2 & PPG & Solar \\
\midrule
DLinear & .181 & .444 & \textbf{.234} & .363 & .150 & .589 & .221 \\
PatchTST & .168 & .478 & .255 & .382 & .243 & .626 & .183 \\
iTransformer & .167 & .474 & .253 & .370 & .184 & .582 & .191 \\
\midrule
Gated KAN (ours) & \textbf{.150} & \textbf{.443} & .245 & \textbf{.343} & \textbf{.147} & \textbf{.565} & \textbf{.182} \\
\bottomrule
\end{tabular}

\vspace{0.5em}
{\footnotesize All models share identical pipeline (RevIN, adaptive norm,
decomposition, patching) and training recipe (Adam, cosine LR, same
epochs/patience/seed=42). Note: results may differ from published numbers
since baselines use our shared preprocessing rather than their original
configurations. The comparison isolates core architecture effects.}
\end{table}

\section{Basis Function Ablation}
\label{app:basis}

\begin{table}[h]
\centering
\caption{Basis function ablation ($H$=96, seed=42). All use the gated
residual architecture; only the KAN basis differs. SinCos achieves
competitive MSE with $\sim$2.5$\times$ fewer parameters than B-spline. Fourier
outperforms B-spline on ETTh2.}
\label{tab:basis}
\small
\begin{tabular}{lcccccc}
\toprule
Basis & Weather & ETTh1 & ETTh2 & Solar & Params \\
\midrule
B-spline (default) & .150 & \textbf{.443} & .245 & \textbf{.181} & 2.19M \\
Fourier (sin/cos) & .153 & .454 & \textbf{.239} & .188 & 2.57M \\
SinCos (learnable $\omega$) & \textbf{.149} & .443 & .239 & --$^*$ & 0.87M \\
MLP (control) & .151 & .444 & .242 & .185 & 0.50M \\
\bottomrule
\end{tabular}
\end{table}

{\footnotesize $^*$SinCos on Solar diverged during training (learnable frequency initialization issue).}

The choice of basis function is dataset-dependent: B-spline is strongest on
Solar (where smooth seasonal splines match the signal), Fourier and SinCos
excel on ETTh2 (periodic structure), and SinCos achieves competitive
performance with the fewest parameters. Temporal Functional Circuits are not
tied to B-splines; any KAN variant with explicit edge functions supports the
framework.

\section{Gate Ablation}
\label{app:gate_ablation}

\begin{table}[h]
\centering
\caption{Gate configuration ablation ($H$=96, seed=42). Among learned-gate
configurations, $\lambda_g=0.05$--$0.10$ gives the best MSE on Weather and
ETTh1; KAN-only remains slightly better on Weather. Without L1, the gate
drifts to $\sim$0.7 (lazy blending).}
\label{tab:gate_ablation}
\small
\begin{tabular}{lcccccc}
\toprule
Config & \multicolumn{2}{c}{Weather} & \multicolumn{2}{c}{ETTh1} & \multicolumn{2}{c}{ETTh2} \\
\cmidrule(lr){2-3}\cmidrule(lr){4-5}\cmidrule(lr){6-7}
 & MSE & $\Ukan$ & MSE & $\Ukan$ & MSE & $\Ukan$ \\
\midrule
KAN-only (no gate) & .149 & -- & .452 & -- & .243 & -- \\
Ungated (gate=1) & .151 & -- & .444 & -- & .245 & -- \\
Fixed gate=0.5 & .153 & 0.50 & .449 & 0.50 & .256 & 0.50 \\
Learned, $\lambda_g$=0 & .153 & 0.68 & .446 & 0.72 & \textbf{.236} & 0.72 \\
Learned, $\lambda_g$=0.01 & .152 & 0.16 & .451 & 0.11 & .240 & 0.16 \\
Learned, $\lambda_g$=0.05 & .150 & 0.05 & \textbf{.443} & 0.02 & .245 & 0.04 \\
Learned, $\lambda_g$=0.10 & \textbf{.150} & 0.02 & .445 & 0.01 & .244 & 0.02 \\
\bottomrule
\end{tabular}
\end{table}

On Weather and ETTh1, stronger L1 ($\lambda_g \geq 0.05$) produces the best
MSE by forcing the gate to be selective. On ETTh2, no regularization
($\lambda_g=0$) works best, allowing the gate to blend freely. Solar
uses $\lambda_g = 0.01$ (selected via validation on this table's
protocol; $\lambda_g = 0.05$ over-suppresses the gate on Solar's
137 variates, reducing $\Ukan$ to $<0.01$). This
dataset-dependent behavior motivates adaptive $\lambda_g$ selection as
future work.

\section{Hyperparameters}
\label{app:hyperparams}

\begin{table}[h]
\centering
\caption{Full hyperparameter specification.}
\small
\begin{tabular}{ll}
\toprule
Hyperparameter & Value \\
\midrule
Input length $L$ & 336 (512 for ETTm1, ECL) \\
Forecast horizon $H$ & 96, 192, 336, 720 \\
Patch length / stride & 16 / 8 \\
Embedding dim $d$ & 32 \\
KAN grid / spline order & 5 / 3 \\
Hidden dim / depth & 64 / 2 \\
Batch size & 256 (Solar: 512, ECL: 128) \\
Optimizer & Adam \\
Initial learning rate & 1e-3 (dataset-specific; see code) \\
Scheduler & Cosine with warmup \\
Epochs / patience & 50 / 10 \\
Gate penalty $\lambda_g$ & 0.05 (Solar: 0.01) \\
Seeds & 42, 123, 456 \\
Train/val/test split & Chronological per LTSF protocol \\
\bottomrule
\end{tabular}
\end{table}

\section{Temporal Functional Circuit Examples}
\label{app:edge_vis}

\begin{figure}[H]
\centering
\includegraphics[width=\textwidth]{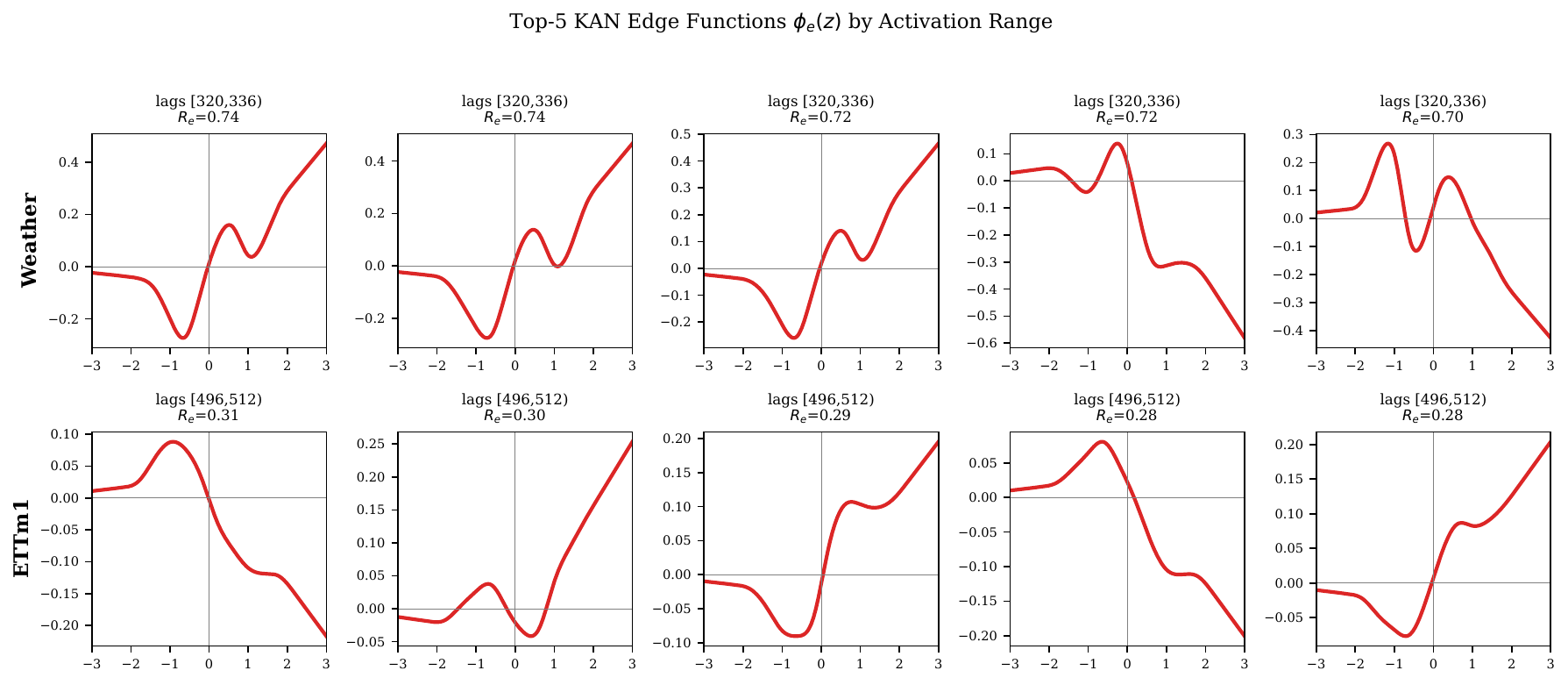}
\caption{Top-5 learned KAN edge functions $\phi_e(z)$ ranked by $R_e$
for the residual branch. Weather edges (top) show complex nonlinear shapes
concentrated on the most recent patch (lags 320--336). ETTm1 edges
(bottom) are smoother with lower $R_e$, consistent with lower $\Ukan$.}
\label{fig:edge_functions}
\end{figure}

Table~\ref{tab:tfc_examples} shows complete Temporal Functional Circuit
tuples $(\phi_e, R_e, \Delta_e, \text{lags})$ for the top-3 Weather
residual edges. All three edges map to the most recent patch (lags 320--336) and have
complex nonlinear shapes (Figure~\ref{fig:edge_functions}). Individual
$\Delta_e$ values are near zero, consistent with the small per-edge
effects observed in Table~\ref{tab:faithful}.

\begin{table}[h]
\centering
\caption{Compact Temporal Functional Circuit summaries for Weather top-3
residual edges. $R_e$: activation range. $I_e$ rank: position in
data-weighted importance ranking (out of 200 scored edges).
$\Delta_e^{\text{zero}}$/$\Delta_e^{\text{spline}}$: MSE change under
zeroing/spline removal.}
\label{tab:tfc_examples}
\small
\begin{tabular}{ccccccc}
\toprule
Edge $(i{\to}j)$ & Lags & $R_e$ & $I_e$ rank & $\Delta_e^{\text{zero}}$ & $\Delta_e^{\text{spline}}$ & Shape \\
\midrule
$1299{\to}17$ & [320, 336) & 0.74 & ${>}200$ & $+6.0 \times 10^{-8}$ & $-6.3 \times 10^{-7}$ & S-curve \\
$1299{\to}8$ & [320, 336) & 0.74 & ${>}200$ & $-2.8 \times 10^{-7}$ & $-1.2 \times 10^{-6}$ & Oscillatory \\
$1308{\to}2$ & [320, 336) & 0.72 & ${>}200$ & $+1.4 \times 10^{-5}$ & $+3.2 \times 10^{-6}$ & Kink \\
\bottomrule
\end{tabular}
\end{table}

\section{Temporal Grounding Validation}
\label{app:temporal_grounding}

\paragraph{Synthetic lag recovery and $A_e$ validation.}
A synthetic signal $x(t) = 0.6\, x(t{-}3) + 0.25\, x(t{-}12) + \epsilon$
($\epsilon \sim \mathcal{N}(0, 0.1^2)$) is
constructed with two known causal lags $\{3, 12\}$ (positions 93 and 84
in a length-96 input window). After training a gated
KAN ($H=16$), multiple attribution methods are used to rank input
positions. This directly validates whether the proposed edge-to-lag
attribution $A_e$ recovers known causal structure.

\begin{center}
\small
\begin{tabular}{lcccc}
\toprule
Method & Recall@5 & Recall@10 & Recall@20 & AUPRC \\
\midrule
$A_e$ (branch-level) & 0.50 & 0.50 & 0.67 & 0.44 \\
Integrated Gradients & 0.50 & \textbf{1.00} & \textbf{1.00} & \textbf{0.53} \\
Gradient $\times$ Input & 0.50 & \textbf{1.00} & \textbf{1.00} & \textbf{0.53} \\
Random & 0.17 & 0.17 & 0.33 & 0.07 \\
\bottomrule
\end{tabular}
\end{center}

$A_e$ (aggregated across all KAN edges via the gradient of the gated
branch output) recovers true lags with AUPRC $6\times$ above random
(0.44 vs.\ 0.07). On the best seed, $A_e$ achieves AUPRC = 0.75 with
both true lags in the top-5. IG achieves higher mean recall by operating
on the full model (both linear and KAN branches jointly), but cannot
identify \emph{which KAN edges} implement the lag dependency---this
edge-level granularity is the unique contribution of TFC's $A_e$.

\paragraph{Attribution stability.}
IG-based temporal attributions on Weather are compared across three
independently trained models (seeds 42, 123, 456). Pairwise Pearson
correlations are $r = 0.94$--$0.97$ and Spearman $\rho = 0.57$--$0.63$.
The same temporal positions are consistently identified as important
regardless of training initialization.

\paragraph{Input-level attribution baselines.}
Masking the top-$k$ attributed input positions and measuring MSE increase
on Weather (Table~\ref{tab:input_masking}):

\begin{table}[h]
\centering
\caption{Input masking on Weather ($H$=96). Relative MSE increase when
masking top-$k$ positions ranked by each method.}
\label{tab:input_masking}
\small
\begin{tabular}{lccc}
\toprule
$k$ & IG & Grad Saliency & Random \\
\midrule
5 & +104\% & +88\% & +3\% \\
10 & +114\% & +103\% & +4\% \\
50 & +126\% & +122\% & +16\% \\
\bottomrule
\end{tabular}
\end{table}

IG and gradient saliency both produce large effects, suggesting the model
relies on specific temporal lags. Random masking produces much smaller
effects, validating that the attribution identifies genuinely important
positions.

\paragraph{$R_e$ vs.\ $I_e$ edge ranking comparison.}
Edge-level deletion curves are computed using three ranking methods:
activation range $R_e$, data-weighted importance $I_e$ (with top-200
prefilter), and random (20 draws). Results span three datasets with
varying $\Ukan$ levels.

\begin{center}
\small
\begin{tabular}{lcrrr}
\toprule
Dataset & $k$ & $\Delta_{R_e}$ & $\Delta_{I_e}$ & $\Delta_{\text{rand}}$ \\
\midrule
\multirow{4}{*}{Weather}
& 5  & $+1.4 \times 10^{-5}$ & $+5.5 \times 10^{-7}$ & $-2.1 \times 10^{-7}$ \\
& 10 & $+4.4 \times 10^{-5}$ & $+1.0 \times 10^{-6}$ & $-1.2 \times 10^{-7}$ \\
& 20 & $+9.0 \times 10^{-5}$ & $+3.4 \times 10^{-6}$ & $-9.0 \times 10^{-7}$ \\
& 50 & $+2.2 \times 10^{-4}$ & $+1.1 \times 10^{-5}$ & $-5.1 \times 10^{-7}$ \\
\midrule
\multirow{2}{*}{ETTm1}
& 10 & $+6.0 \times 10^{-7}$ & $-3.0 \times 10^{-8}$ & $-4.8 \times 10^{-8}$ \\
& 50 & $+2.1 \times 10^{-6}$ & $+6.0 \times 10^{-8}$ & $-2.2 \times 10^{-8}$ \\
\midrule
\multirow{2}{*}{ETTh1}
& 10 & $-1.2 \times 10^{-7}$ & $+3.0 \times 10^{-8}$ & $+3.0 \times 10^{-9}$ \\
& 50 & $-4.8 \times 10^{-7}$ & $-8.9 \times 10^{-8}$ & $-6.4 \times 10^{-8}$ \\
\bottomrule
\end{tabular}
\end{center}

On Weather ($\Ukan = 0.046$), $R_e$ produces 20--42$\times$ larger
deletion effects than $I_e$ across all $k$ values, and the effect grows
monotonically with $k$. On ETTm1 ($\Ukan = 0.041$), $R_e$ still
dominates but absolute effects are smaller. On ETTh1
($\Ukan = 0.022$), all three methods produce near-zero effects
indistinguishable from noise, consistent with the gate suppressing the
KAN branch. This pattern provides evidence that $R_e$ faithfully
identifies important edges when the KAN branch is active, and that the
gating mechanism correctly governs when edge-level analysis is
informative. A combined metric $S_e = R_e \cdot I_e$ was also tested but
produced identical deletion effects to $I_e$ alone, suggesting the two
metrics capture overlapping information. For B-spline KANs, learned
functional capacity (activation range) is a strong and computationally
efficient faithfulness signal.

\paragraph{Multi-seed deletion with statistical tests.}
Deletion experiments are repeated across 3 training seeds (42, 123, 456)
with 50 random draws per seed and 1000-sample bootstrap confidence intervals.

\begin{center}
\small
\begin{tabular}{lcrrcr}
\toprule
Dataset & $k$ & $\Delta_{\text{top}}$ (mean) & 95\% CI & $p$ & RelDeg \\
\midrule
\multirow{2}{*}{Weather}
& 10 & $+4.6 \times 10^{-5}$ & $[3.8, 5.0] \times 10^{-5}$ & ${<}10^{-4}$ & 0.031\% \\
& 50 & $+1.8 \times 10^{-4}$ & $[1.7, 2.1] \times 10^{-4}$ & ${<}10^{-4}$ & 0.123\% \\
\midrule
\multirow{2}{*}{ETTm2}
& 10 & $+8.9 \times 10^{-6}$ & --- & ${<}10^{-3}$ & 0.006\% \\
& 50 & $+3.3 \times 10^{-5}$ & --- & ${<}10^{-4}$ & 0.022\% \\
\midrule
\multirow{2}{*}{Solar}
& 10 & $+3.6 \times 10^{-5}$ & --- & ${<}10^{-4}$ & 0.020\% \\
& 50 & $+1.4 \times 10^{-4}$ & --- & ${<}10^{-4}$ & 0.079\% \\
\bottomrule
\end{tabular}
\end{center}

Top-$k$ deletion effects are statistically significant ($p < 10^{-3}$)
across all seeds on Weather, ETTm2, and Solar. PPG shows non-significant
effects ($p > 0.3$), consistent with $\Ukan \approx 0.02$.
RelDeg = $\Delta / \text{MSE}_{\text{base}} \times 100$.

\paragraph{Normalized KAN-branch deletion.}
To contextualize the small absolute deletion effects, the full
KAN-branch removal effect $\Delta_{\text{all-KAN}}$ is measured (zeroing
all KAN layers in both branches):

\begin{center}
\small
\begin{tabular}{lrrrrr}
\toprule
Dataset & $\Delta_{\text{all-KAN}}$ & RelDeg & $\eta_{50}$ \\
\midrule
Weather & $+1.03 \times 10^{-2}$ & 6.9\% & 2.1\% \\
Solar & $+9.04 \times 10^{-2}$ & 49.7\% & 0.16\% \\
\bottomrule
\end{tabular}
\end{center}

The KAN branch contributes 6.9\% on Weather and 49.7\% on Solar.
Individual edge effects are small because the contribution is distributed
across ${\sim}$22K edges; the top-50 edges capture $\eta_{50} = \Delta_{\text{top-50}} / \Delta_{\text{all-KAN}}$ of the total
branch effect.

\paragraph{$R_{\text{KAN}}$ on synthetic regimes.}
The calibrated contribution metric
$R_{\text{KAN}} = \E[\|g(x) \fkan(x)\|_2 / \|\yhat(x)\|_2]$
shows stronger discrimination than $\Ukan$ on synthetic regimes
($\lambda_g = 0$):

\begin{center}
\small
\begin{tabular}{lcc}
\toprule
Regime & $\Ukan$ & $R_{\text{KAN}}$ \\
\midrule
Linear (sine) & 0.49 & 0.046 \\
Multi-freq periodic & 0.53 & 0.127 \\
Threshold AR & 0.56 & 0.497 \\
Regime-switching & 0.72 & 0.420 \\
\bottomrule
\end{tabular}
\end{center}

While $\Ukan$ already reaches 0.49 on the linear regime (the gate
has no sparsity incentive to close), $R_{\text{KAN}}$ correctly shows
that the KAN branch contributes only 4.6\% of the forecast on linear
data vs.\ 50\% on the nonlinear threshold AR.

\paragraph{Faithfulness sanity checks.}
Following \citet{adebayo2018sanity}, deletion ranking is tested after
randomizing model weights. On Weather:

\begin{center}
\small
\begin{tabular}{lc}
\toprule
Condition & Top-50 $\Delta$MSE \\
\midrule
Trained model (normal) & $+2.20 \times 10^{-4}$ \\
Randomized all KAN weights & $0.0$ \\
Randomized spline coefficients only & $+7.4 \times 10^{-5}$ \\
\bottomrule
\end{tabular}
\end{center}

Randomizing all KAN weights eliminates the deletion effect entirely,
confirming that the ranking reflects learned structure. Randomizing only
spline coefficients (keeping base SiLU weights) reduces but does not
eliminate the effect, consistent with the base activation carrying partial
signal.

\paragraph{Extended deletion: fraction of KAN branch captured.}
To address whether the nonlinear correction is concentrated or diffuse,
the fraction $\eta_k = \Delta_{\text{top-}k} / \Delta_{\text{all-KAN}}$
is computed over a wider range of $k$:

\begin{center}
\small
\begin{tabular}{lrrrrrrr}
\toprule
& \multicolumn{7}{c}{$\eta_k$ (\% of KAN branch effect)} \\
Dataset & $k$=50 & 100 & 250 & 500 & 1K & 2.5K & 5K \\
\midrule
Weather & 2.1 & 3.7 & 8.1 & 13.1 & 20.0 & 29.2 & 35.0 \\
Solar & 0.2 & 0.3 & 0.6 & 1.1 & 2.2 & 4.1 & 7.8 \\
\bottomrule
\end{tabular}
\end{center}

On Weather ($\Delta_{\text{all-KAN}} = +0.010$, 6.9\% relative), the
top 1\% of edges (1K of ${\sim}$84K) capture 20\% of the branch effect,
and the top 6\% capture 35\%. The nonlinear correction is distributed
but not uniform: a small fraction of edges carries disproportionate load.
Solar's KAN branch is larger ($\Delta_{\text{all-KAN}} = +0.090$, 50\%)
but more diffuse across its 137 variates.

\paragraph{Gate sweep: $R_{\text{KAN}}$ under sparsity pressure.}
With $\lambda_g > 0$, the gate closes on linear data but remains open
on nonlinear regimes:

\begin{center}
\small
\begin{tabular}{llcccc}
\toprule
Regime & $\lambda_g$ & $\Ukan$ & $R_{\text{KAN}}$ & MSE & $\Delta_{\text{all-KAN}}$ \\
\midrule
\multirow{3}{*}{Linear}
& 0 & 0.48 & 0.046 & .0026 & +.0025 \\
& 0.01 & 0.00 & 0.000 & .0026 & 0 \\
& 0.05 & 0.00 & 0.000 & .0026 & 0 \\
\midrule
\multirow{3}{*}{Threshold AR}
& 0 & 0.55 & 0.475 & .951 & +.0004 \\
& 0.01 & 0.14 & 0.263 & .952 & +.0004 \\
& 0.05 & 0.02 & 0.080 & .953 & $-$.0001 \\
\midrule
\multirow{3}{*}{Regime-switch}
& 0 & 0.72 & 0.373 & .011 & +.143 \\
& 0.01 & 0.06 & 0.141 & .011 & +.056 \\
& 0.05 & 0.00 & 0.023 & .019 & +.013 \\
\bottomrule
\end{tabular}
\end{center}

Any $\lambda_g > 0$ fully closes the gate on linear data ($R_{\text{KAN}} = 0$,
$\Delta_{\text{all-KAN}} = 0$). On regime-switching, the KAN branch
remains essential even at $\lambda_g = 0.05$: removing it increases MSE
by $+0.013$ (from 0.019 to 0.032), and $R_{\text{KAN}} = 0.023$ indicates
residual nonlinear contribution. At $\lambda_g = 0$, the KAN branch
accounts for $\Delta_{\text{all-KAN}} = +0.143$, a 13$\times$ increase
in MSE.

\clearpage


\newpage
\section*{NeurIPS Paper Checklist}

\begin{enumerate}

\item {\bf Claims}
    \item[] Question: Do the main claims made in the abstract and introduction accurately reflect the paper's contributions and scope?
    \item[] Answer: \answerYes{}
    \item[] Justification: The abstract states that no single core dominates, that gate utilization tracks signal complexity on synthetic data, and that edge deletion curves show faithfulness. All claims are supported by the reported tables and figures.

\item {\bf Limitations}
    \item[] Question: Does the paper discuss the limitations of the work performed by the authors?
    \item[] Answer: \answerYes{}
    \item[] Justification: Section~7 discusses five explicit limitations: model-internal (not real-world causal) interventions, gradient saturation in deep layers, KAN edges are not automatically human-semantic, performance is competitive but not uniformly SOTA, and synthetic regime discrimination is imperfect.

\item {\bf Theory assumptions and proofs}
    \item[] Question: For each theoretical result, does the paper provide the full set of assumptions and a complete (correct) proof?
    \item[] Answer: \answerNA{}
    \item[] Justification: The paper does not make theoretical claims. It is an empirical XAI framework paper.

\item {\bf Experimental result reproducibility}
    \item[] Question: Does the paper fully disclose all the information needed to reproduce the main experimental results?
    \item[] Answer: \answerYes{}
    \item[] Justification: Section~5.1 and Appendix~F specify all architecture details, training hyperparameters, dataset splits, and random seeds. Anonymized source code is included in the supplementary material.

\item {\bf Open access to data and code}
    \item[] Question: Does the paper provide open access to the data and code?
    \item[] Answer: \answerPartly{}
    \item[] Justification: All datasets are publicly available standard benchmarks. PPG-DaLiA is from the UCI Repository under CC BY 4.0. Anonymized source code is included in the supplementary material.

\item {\bf Experimental setting/details}
    \item[] Question: Does the paper specify all the training and test details?
    \item[] Answer: \answerYes{}
    \item[] Justification: Section~5.1 specifies optimizer, learning rate schedule, epochs, patience, gate regularization, patch size, stride, KAN grid/order, and batch sizes.

\item {\bf Experiment statistical significance}
    \item[] Question: Does the paper report error bars suitably and correctly?
    \item[] Answer: \answerYes{}
    \item[] Justification: Table~1 reports mean $\pm$ std over 3 seeds. Synthetic experiments (Table~3) also report 3-seed statistics. ETTh1 shows high seed variance; other datasets are stable.

\item {\bf Experiments compute resources}
    \item[] Question: Does the paper provide sufficient information on compute resources?
    \item[] Answer: \answerYes{}
    \item[] Justification: Experiments were run on NVIDIA H100 80GB GPUs. The model has 2.2M parameters and trains in 2--30 minutes per dataset depending on variate count.

\item {\bf Code of ethics}
    \item[] Question: Does the research conform with the NeurIPS Code of Ethics?
    \item[] Answer: \answerYes{}
    \item[] Justification: The work uses publicly available benchmarks. PPG-DaLiA is a previously published, anonymized public dataset. No private data or dual-use concerns.

\item {\bf Broader impacts}
    \item[] Question: Does the paper discuss potential societal impacts?
    \item[] Answer: \answerPartly{}
    \item[] Justification: Faithful temporal explanations could benefit domains requiring transparent forecasts (healthcare, energy). Interpretability claims are carefully qualified to avoid overstating readiness for safety-critical deployment.

\item {\bf Safeguards}
    \item[] Question: Does the paper describe safeguards for responsible release?
    \item[] Answer: \answerNA{}
    \item[] Justification: Standard forecasting architecture with public benchmarks. No safeguards needed beyond standard open-source practices.

\item {\bf Licenses for existing assets}
    \item[] Question: Are creators of assets properly credited?
    \item[] Answer: \answerYes{}
    \item[] Justification: All datasets and libraries are cited. PPG-DaLiA under CC BY 4.0. efficient-kan library credited.

\item {\bf New assets}
    \item[] Question: Are new assets well documented?
    \item[] Answer: \answerYes{}
    \item[] Justification: The Gated Residual KAN codebase, XAI utilities, and experiment scripts will be released with documentation.

\item {\bf Crowdsourcing and human subjects}
    \item[] Question: For research with human subjects, does the paper include instructions given to participants?
    \item[] Answer: \answerNA{}
    \item[] Justification: No new human-subject data collected.

\item {\bf IRB approvals}
    \item[] Question: Does the paper describe potential risks to study participants?
    \item[] Answer: \answerNA{}
    \item[] Justification: No new human-subject data collected.

\item {\bf Declaration of LLM usage}
    \item[] Question: Does the paper describe the usage of LLMs?
    \item[] Answer: \answerYes{}
    \item[] Justification: Large language models were used during manuscript preparation for prose editing and code generation assistance. All scientific content, experimental design, architecture design, and analysis are the original work of the authors.

\end{enumerate}

\end{document}